\documentclass[final,3p,times]{elsarticle}

\usepackage{amssymb}
\usepackage{amsmath}
\usepackage{multirow, multicol}
\usepackage{booktabs}
\usepackage{algorithmic}
\usepackage{graphicx}
\usepackage{textcomp}
\usepackage{pifont}
\usepackage{amsfonts}
\usepackage[normalem]{ulem}
\newcommand{\cmark}{\ding{51}}%
\newcommand{\xmark}{\ding{55}}%

\begin{document}

\begin{frontmatter}



\title{Rethinking Text-Promptable Surgical Instrument Segmentation with Robust Framework}


\author[label1]{Tae-Min Choi}
\author[label1]{Juyoun Park} 

\affiliation[label1]{organization={Center for humanoid research, Korea Institute of Science Technology (KIST)},
            addressline={5, Hwarang-ro 14-gil Seongbuk-gu}, 
            city={Seoul},
            postcode={02792}, 
            country={South Korea}}

\begin{abstract}
Surgical instrument segmentation is an essential component of computer-assisted and robotic surgery systems. Vision-based segmentation models typically produce outputs limited to a predefined set of instrument categories, which restricts their applicability in interactive systems and robotic task automation. Promptable segmentation methods allow selective predictions based on textual prompts. However, they often rely on the assumption that the instruments present in the scene are already known, and prompts are generated accordingly, limiting their ability to generalize to unseen or dynamically emerging instruments. In practical surgical environments, where instrument existence information is not provided, this assumption does not hold consistently, resulting in false-positive segmentation. To address these limitations, we formulate a new task called Robust text-promptable Surgical Instrument Segmentation (R-SIS). Under this setting, prompts are issued for all candidate categories without access to instrument presence information. R-SIS requires distinguishing which prompts refer to visible instruments and generating masks only when such instruments are explicitly present in the scene. This setting reflects practical conditions where uncertainty in instrument presence is inherent. We evaluate existing segmentation methods under the R-SIS protocol using surgical video datasets and observe substantial false-positive predictions in the absence of ground-truth instruments. These findings demonstrate a mismatch between current evaluation protocols and real-world use cases, and support the need for benchmarks that explicitly account for prompt uncertainty and instrument absence.
\end{abstract}





\end{frontmatter}



\section{Introduction}
\label{sec:introduction}
Minimally invasive surgeries \cite{fuchs2002minimally, harrell2005minimally} are widely adopted due to their advantages, requiring surgeons to rely on camera-captured views instead of direct visualization. Surgical Instrument Segmentation (SIS) \cite{shvets2018automatic} is fundamental to enhancing surgical vision. Vision-based SIS models \cite{iglovikov2018ternausnet, jin2019incorporating, zhao2020learning, gonzalez2020isinet, baby2023forks, ayobi2023matis} typically generate predictions for all predefined instrument categories, regardless of whether they are present in the surgical scene. While effective in static settings, this behavior limits integration with interactive systems and automation pipelines, where selective and context-aware segmentation is often required. More recently, promptable segmentation approaches \cite{zhou2023text, yue2023part} have been introduced, allowing segmentation to be guided by natural language descriptions. Many recent promptable segmentation approaches are inspired by referring image segmentation (RIS) \cite{yang2022lavt, yu2018mattnet}, which assumes that a query refers to an object that is present in the image. RIS methods offer selective segmentation over target categories but commonly rely on the assumption that the prompted object is visible in the image. This assumption in RIS also applies to the promptable SIS method.

\begin{figure}
    \centering
    \includegraphics[width=\textwidth]{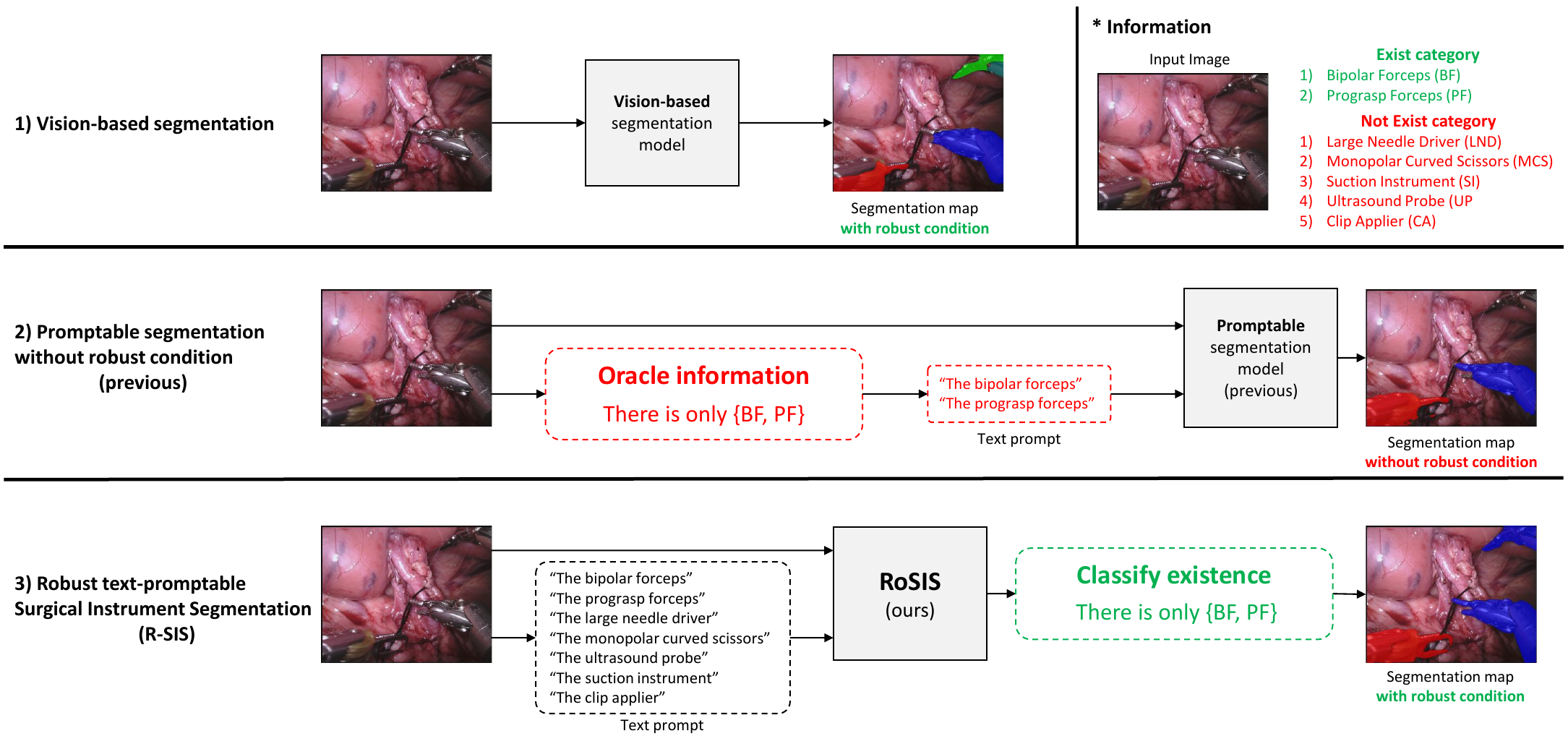}
    \caption{Comparison of evaluation process of SIS method: (1) Vision-based segmentation model \cite{iglovikov2018ternausnet, jin2019incorporating, zhao2020learning, gonzalez2020isinet} only uses the image and generates all classes of segmentation maps; (2) Previous promptable segmentation model \cite{zhou2023text, yue2024surgicalsam, yue2023part} uses ``oracle information" to identify categories present in the image and generates segmentation maps only for those categories. This means they are led to settings without robust conditions; (3) Our R-SIS uses prompts for all classes, generating segmentation maps for all categories under robust conditions.}
    \label{fig:fig1}
\end{figure}

However, this assumption does not consistently hold in real surgical environments, where we do not know the presence of instruments in the current scene (oracle information). This leads promptable segmentation models to often produce false-positive masks for absent instruments when prompted without presence verification \cite{wu2024towards}. We need a stricter standard in surgical environments due to the critical risks posed by false-positive masks when instrument presence is uncertain. Moreover, current evaluation protocols for promptable segmentation methods often rely on oracle information that is unavailable in practical scenarios. As illustrated in Fig. \ref{fig:fig1}, these models \cite{zhou2023text, yue2024surgicalsam, yue2023part} exclude prompts for absent instruments based on the ground truth, leading to evaluations that do not reflect real-world surgical environments. 




To address these limitations, we introduce a new task formulation: \textit{Robust text-promptable Surgical Instrument Segmentation (R-SIS)}. Under this setting, prompts are issued for all candidate categories without assuming knowledge of instrument presence. A model must determine whether the prompted instrument is visually present in the current scene and, only if confirmed, produce an appropriate segmentation mask. This process ensures that segmentation is based on visual features rather than relying on prior assumptions about instrument presence. This setting is also fit for real surgical systems, where the presence of instruments changes frequently in the camera's view.

To support this task, we design a segmentation framework, referred to as \textit{Robust Surgical Instrument Segmentation (RoSIS)}, that integrates visual and textual features, predicts instrument existence, and generates segmentation outputs using multiple prompt types. The framework supports diverse prompt types and introduces iterative refinement to achieve improved segmentation performance under the R-SIS setting, without relying on prior knowledge of instrument presence.

To evaluate segmentation performance under the R-SIS setting, we conduct experiments on two benchmark datasets, EndoVis2017 \cite{allan20192017} and EndoVis2018 \cite{allan20202018}. Through a series of detailed and diverse experiments, we demonstrate how prompt uncertainty and instrument absence affect segmentation behavior and highlight the importance of realistic evaluation protocols.

In summary, our contributions are three-fold: (1) We define the R-SIS task, which evaluates promptable segmentation under realistic conditions without assuming instrument presence. (2) We present the RoSIS, which is a segmentation framework that supports existence prediction and iterative refinement using diverse prompt types. (3) We establish R-SIS evaluation protocols and benchmarks on two surgical datasets, revealing limitations in existing methods and motivating more robust prompting strategies.


\begin{figure}[t]
    \centering
    \includegraphics[width=\textwidth]{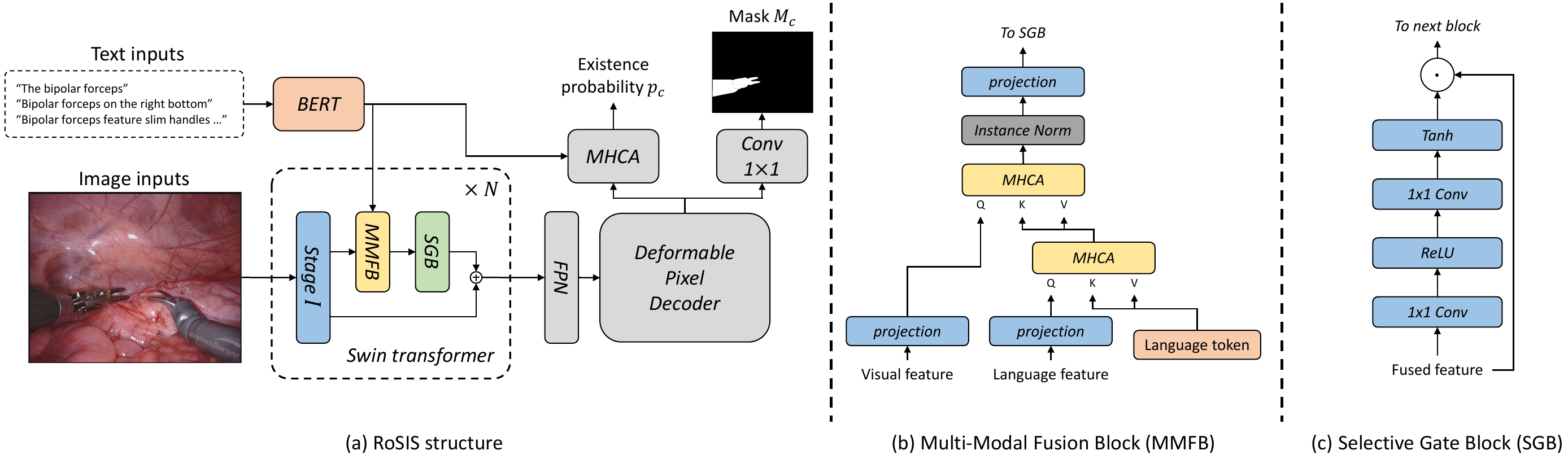}
    \caption{RoSIS model architecture with (a) overall structure, (b) Multi-Modal Fusion Block, and (c) Selective Gate Block.}
    \label{fig:fig3}
\end{figure}

\section{Related Works}
\subsection{Surgical Instrument Segmentation}
SIS involves predicting the pixel-wise regions of instruments in surgical images. Early research \cite{iglovikov2018ternausnet, gonzalez2020isinet, zhao2022trasetr, ayobi2023matis, baby2023forks} applied image-based segmentation methods to SIS. TernausNet \cite{iglovikov2018ternausnet} introduced a modified U-Net \cite{ronneberger2015u} architecture with pre-trained VGG11/VGG16 encoders and skip connections for precise location. ISINet \cite{gonzalez2020isinet}, utilizing a Mask R-CNN \cite{he2017mask} architecture, incorporates a temporal consistency module to refine instrument predictions across video frames. TraSeTR \cite{zhao2022trasetr} employs a Track-to-Segment Transformer, combining query embeddings with contrastive learning for robust tracking and segmentation. MATIS \cite{ayobi2023matis}, a video-based SIS method, uses a two-stage structure with Mask2Former \cite{cheng2022masked} to generate and classify region proposals, incorporating masked and deformable attention and video transformers for temporal consistency. S3Net \cite{baby2023forks} extends Mask R-CNN \cite{he2017mask} with a Multi-Scale Mask-Attended classifier, refining class predictions based on mask-attended features to improve instance segmentation. Recently, with advancements in segmentation techniques utilizing information beyond vision, promptable segmentation models incorporating external prompts, such as text or prototypes, have been proposed for SIS.

SurgicalSAM \cite{yue2024surgicalsam} introduces an efficient fine-tuning method for the Segment Anything Model (SAM) \cite{kirillov2023segment}, which employs prompts (points, bounding boxes, or text inputs) from a prompt encoder. Since SAM was originally designed for natural domain image segmentation, SurgicalSAM adapts it for the surgical domain. Additionally, it introduces a prototype-based class prompt encoder that generates prompt embeddings for each class as inputs to SAM. Later, various text-promptable SIS methods were proposed, leveraging vision-language models for text-based segmentation. SP-SAM \cite{yue2023part} builds upon SAM \cite{kirillov2023segment}, incorporating both category-level and part-level text prompts. TP-SIS \cite{zhou2023text}, based on CLIP \cite{radford2021learning}, utilizes text prompts describing instruments, categorized into name-based prompts, CLIP’s prompt templates, and prompts generated from GPT-4 \cite{achiam2023gpt}. However, these methods assume that the described instrument is always present in the image, leading to incorrect masks when the instrument is absent. To mitigate this, existing approaches first identify the instruments and then generate prompts based on detected instruments. This introduces oracle information during evaluation, leading to unfair comparisons with vision-only models.

We rethink text-promptable SIS by introducing R-SIS, which eliminates this assumption and ensures a fair evaluation. Unlike prior methods that selectively use prompts for detected instruments, R-SIS systematically integrates prompts for all classes, allowing the model to infer instrument presence dynamically. Our approach also addresses the false positives inherent in previous text-promptable SIS models.

\subsection{Referring Image Segmentation}
RIS aims to segment objects based on a given text expression. Early works like MAttNet \cite{yu2018mattnet} used language attention networks and object proposals to match expressions with objects. Later methods, such as VLT \cite{ding2021vision} and EFN \cite{feng2021encoder}, employed Bi-GRU language models and ResNet-based encoders to fuse multi-modal information. The adoption of Vision Transformers (ViT) \cite{dosovitskiy2020image} has led to further advancements in RIS. LAVT \cite{yang2022lavt} integrates BERT-based text encoding with visual features from ViT through pixel-word attention, significantly improving segmentation accuracy.

However, a key limitation of prior RIS models is the assumption that every text expression corresponds to an object in the image. This assumption leads to incorrect mask generation when no relevant object is present. Generalized Referring Expression Segmentation \cite{liu2023gres} tackles this issue by introducing empty-target (no target) and multi-target scenarios. RefSegformer \cite{wu2024towards} further refines this approach by explicitly handling the empty-target problem, ensuring robust segmentation even when the described object is absent. This issue also affects text-promptable SIS methods, where prompts often refer to instruments that do not exist in the image. Our work extends the robust RIS framework to SIS, addressing false positives while ensuring fair comparisons between promptable and vision-based segmentation models.

\section{Method}
This section presents our approach to R-SIS and RoSIS. Section \ref{sec:sec3.1} formulates the problem for R-SIS, identifying the notation and procedure of promptable segmentation for surgical instruments. Section \ref{sec:sec3.2} defines the construction of text prompts used for training, detailing the use of descriptive prompts to enhance model understanding. Section \ref{sec:sec3.3} describes the architecture of RoSIS, which is an encoder-decoder structure with multi-modal fusion components and an instrument existence prediction branch. Finally, Section \ref{sec:sec3.4} introduces an iterative refinement strategy, applied during inference, that is designed to address the R-SIS task by determining the presence of instruments and leveraging diverse prompts to generate refined segmentation masks.
\subsection{Problem formulation}
\label{sec:sec3.1}
SIS is the task of segmenting surgical tools in an endoscopic surgery image \( I \in \mathbb{R}^{H \times W \times 3} \), where \( H \) and \( W \) denote the height and width of the image, respectively. For text-promptable SIS, we incorporate an image \( I \) and a text prompt \( T \) containing the names or attributes of the surgical tools, making the segmentation task responsive to specific prompts.

In traditional vision-based SIS, the model takes only image \( I \) as input and predicts a mask \( M \in \mathbb{R}^{H \times W \times C} \) that includes \( C \) classes, each corresponding to a specific surgical instrument. However, in text-promptable SIS, both the image \( I \) and a class-specific text prompt \( t_c \) are provided as input for each class \( c \). This enables the model to generate a binary mask \( M_c \in \mathbb{R}^{H \times W \times 1} \) for each class \( c \). For the R-SIS, we also generate an existence probability \( p_c \), which classifies the class \( c \) in the image for a given text prompt \( t_c \).

In the test phase, we conduct inference across all \( C \) classes for a given image \( I \), resulting in class-specific masks \( M_1, \ldots, M_C \) and existence probabilities \( p_1, \ldots, p_C \). These outputs are then aggregated to the final mask \( M \in \mathbb{R}^{H \times W \times C} \), with each pixel in \( M \) representing the class-specific segmentation.

To enable R-SIS, we incorporate existence probabilities, allowing our model to assess the presence of each instrument class first. This setup enables to avoidance of producing unnecessary masks for absent classes, reducing false positives, and enhancing segmentation accuracy under the R-SIS setting. By integrating class existence probabilities, our method is evaluated under robust conditions with vision-based models and demonstrates results under practical conditions.

\subsection{Prompt generation}
\label{sec:sec3.2}
There are various ways to create text prompts \( t_c \) for each class \( c \). Following the approach in \cite{zhou2023text}, we generate prompts using the class name, such as ``The \{cls name\}" and we utilize GPT-4 \cite{achiam2023gpt} to create detailed prompts describing each class’s visual characteristics and keep them during the train and test phase. This combination provides both general and specific prompts, enabling the model to recognize instruments based on both name and appearance.

Since surgical instruments often appear in consistent positions on the screen during surgery, we further enhance the prompts by designing location-based descriptions. To do this, we calculate the center of mass for each instrument's ground truth mask to identify its typical position, creating prompts like ``The \{cls name\} on the \{location\}." During training, these location prompts are generated using ground truth annotations to ensure accurate positional information. For location prompts, we categorize positions into four options: ``left-top," ``left-bottom," ``right-top," and ``right-bottom." During model training, we employ all three types of text prompts—class name prompt, GPT-4 visual prompt, and location prompt—for each class \( c \), ensuring comprehensive contextual information. During inference, since ground truth is not accessible, we apply an iterative refinement strategy to dynamically estimate instrument locations and construct corresponding location prompts, thereby maximizing their utility under deployment conditions.

If training only with positive prompts (describing instruments present in the image), the model may develop a bias toward the presence of instruments, potentially reducing generalization \cite{wu2024towards}. To counter this, we also generate negative prompts for classes absent from the image, thus enhancing the classification ability to distinguish between existing and non-existing classes. For absent classes, we create prompts using both the class name and GPT-4 description prompts, as well as a randomly selected location prompt from the four options, to provide a full range of negative examples. We balance the positive and negative prompts corresponding to one image for stable training.

\subsection{Architecture}
\label{sec:sec3.3}
We design our \textit{Robust Surgical Instrument Segmentation (RoSIS)} using an encoder-decoder structure to enable R-SIS. Inspired by \cite{wu2024towards} and \cite{liu2023gres}, we add a binary classifier in parallel with the decoder to determine the existence of instruments. First, we encode the image \( I \) and text description \( T \) using an image encoder and a text encoder, respectively. The image encoder is based on the Swin Transformer \cite{liu2021swin}, which is specialized for dense prediction tasks, while the text encoder uses BERT \cite{devlin2018bert}. We denote the visual feature from stage \( i \) of the Swin Transformer as \( v_i \in \mathbb{R}^{H_i \times W_i \times C_i} \) and the language feature from BERT as \( l \in \mathbb{R}^{T \times C_l} \), where \( H_i \) and \( W_i \) are the height and width of the visual feature, \( T \) is the number of words in the text description, and \( C_i \) and \( C_l \) represent the number of channels for the visual and language features, respectively.

\subsubsection{Encoder design}
We introduce early feature fusion to efficiently integrate vision and language knowledge from the encoder. We aim to enhance semantic understanding and improve feature interactivity through early fusion. For this purpose, we add the \textit{MMFB} and the \textit{SGB} between each stage of the Swin Transformer.

Fig. \ref{fig:fig3}(b) shows the architecture of the MMFB, which is designed to integrate visual and language features through a multi-step fusion process, with the fused feature then passed to a gating module. The fusion process begins by independently projecting the visual and language features into compatible feature spaces. The MMFB contains two Multi-Head Cross Attention (MHCA) modules. In the first MHCA, we adopt language tokens to fuse with the language feature. The language token is a learnable parameter that adjusts the language feature before the fusion. Unlike \cite{wu2024towards}, since we do not use the language features generated during fusion in the decoder, we focus on selecting only the information necessary for fusion with visual features. Therefore, we refine the language features with the first MHCA and subsequently use them in the second MHCA to interact with the visual features.

Inspired by \cite{yang2022lavt}, after the MMFB, we introduce the SGB to regulate the flow of fused features, ensuring that language features do not dominate the visual information. The SGB dynamically controls feature importance, allowing the model to maintain a balanced representation of both visual and language features before passing them to the next stage. As shown in Fig. \ref{fig:fig3}(c), the SGB takes the fused feature as input, processes it through a series of lightweight transformations and applies a gating mechanism.

The whole fusion process between each stage are formulated as follows:
\begin{align}
v_i &= V_i(f_{i-1}),\quad i \in \{1, 2, 3, 4\} \\
f_i &= 
\begin{cases}
    v_i + \text{SGB}_i(\text{MMFB}_i(v_i,\ l)),\ i \in \{1, 2, 3, 4\}, \\
    I,\ i = 0
\end{cases}
\end{align}
where \( V_i \) represents the \( i\text{-th} \) stage of the Swin Transformer, and \( \text{SGB}_i \) and \( \text{MMFB}_i \) denote the \( i\text{-th} \) fusion blocks placed between \( V_i \) and \( V_{i+1} \). We describe the effects of the encoder fusion architecture in the ablation study in Table \ref{tab:tab3}.

\subsubsection{Decoder design}
Our decoder is designed to perform two tasks: classify the existence of the instrument specified by the text prompt and generate the corresponding segmentation mask. To accomplish this, we use a multi-scale deformable attention pixel decoder \cite{zhu2020deformable}. This decoder incorporates an Feature Pyramid Networks \cite{lin2017feature} structure to effectively utilize multi-scale features, taking \( f_1, f_2, f_3, \) and \( f_4 \) as inputs. The final segmentation mask \(M_c\) is produced to pass the output from the decoder through a \( 1 \times 1 \) convolution layer.

In addition to mask generation, we introduce a parallel branch to compute the existence probability of the instrument described by the text prompt. This branch fuses the decoder output from \( f_4 \) with the raw language feature extracted from BERT using MHCA. This fusion enables the decoder to leverage both visual and textual information, enhancing its ability to determine the presence of the specified instrument. Unlike \cite{wu2024towards}, which uses fused language features from the encoder for existence prediction, we use raw BERT embeddings as input language features to preserve rich semantic context without introducing task-specific projection layers. While prior work adopts cross-modal embeddings aligned during training, such methods often entangle vision and language representations, which may bias prompt behavior in the presence of instrument absence.
After the fusion through MHCA, the existence probability \( p_c\) is computed by passing the output through a linear layer.

Finally, we calculate the cross-entropy loss for both the generated mask and the existence probability to optimize the model’s accuracy in both segmentation and existence detection tasks, shown as follows:
\begin{equation}
    L = BCELoss(p_c, y_c) + \lambda \cdot CELoss(M_c, M_c^{gt})
\end{equation}
where \(\lambda\) is a hyperparameter for mask loss, \(y_c\) and \(M_c^{gt}\) denote the ground truth of the existence and the mask of class \(c\), respectively.
\begin{figure}[t]
    \centering
    \includegraphics[width=\textwidth]{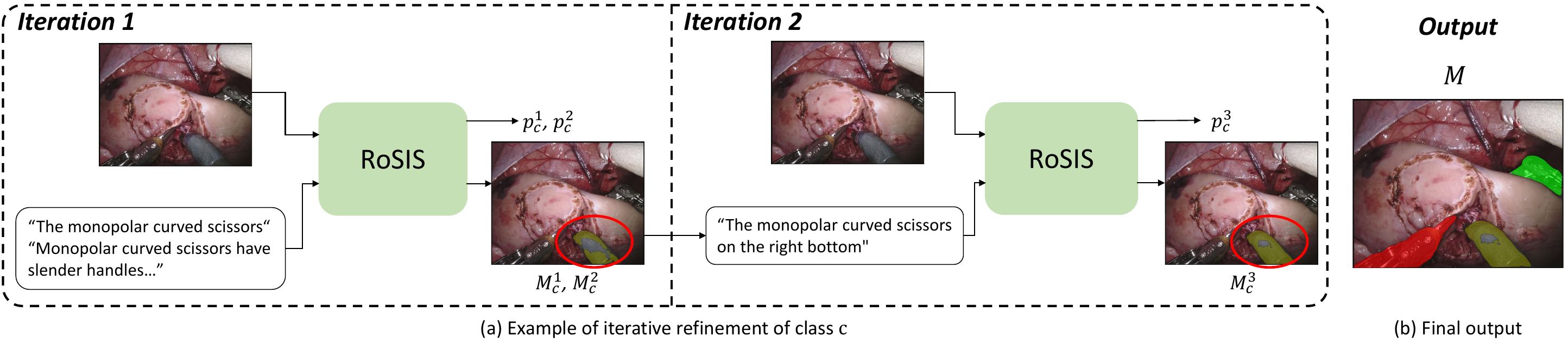}
    \caption{(a) Iterative refinement process of class $c$ (monopolar curved scissors in this image): (Iteration 1) Initial segmentation using a general prompt, (Iteration 2) refined segmentation with a location prompt referring to the first predicted map. Then, we combine maps by equation \ref{eq:eq3} for accurate final output. (b) After applying the iterative refinement process to all classes, the refined maps are combined to generate the final output.}
    \label{fig:fig4}
\end{figure}
\subsection{Iterative refinement for inference}
\label{sec:sec3.4}
We propose an iterative refinement strategy to use location prompts during inference. During training, given the ground-truth instruments present in the scene, RoSIS learns to predict segmentation masks by first determining the presence of instruments and utilizing diverse prompts, thereby reducing false positives under robust settings. During inference, where instrument existence information is not available, the model performs a two-iteration inference process, as illustrated in Fig. \ref{fig:fig4}: it first predicts which instruments are visible by applying all instrument prompts, and then generates refined masks only for those instruments that are visually present in the scene.

In the first iteration, we generate the initial segmentation map and existence probability for each instrument category. We use two prompts: a category name prompt (e.g., “The monopolar curved scissors”) and a descriptive prompt generated by GPT-4, which provides visual characteristics of the instrument. For category \(c\), the model predicts the segmentation maps \(M_c^1 \text{ and } M_c^2\) and the existence probabilities \(p_c^1 \text{ and } p_c^2\). If the average existence probability \((p_c^1+p_c^2) / 2\) exceeds 0.5, indicating that class \(c\) exists in the image, the process proceeds to the second iteration.

In the second iteration, while location information can be directly obtained from the ground truth during training, it is not available during inference. Therefore, the model instead leverages the predicted masks from the first iteration to guide refinement. Based on the segmentation maps \(M_c^1 \text{ and } M_c^2\) from the first iteration, the model calculates the center of mass for the detected instrument, allowing it to determine a rough location within one of four quadrants: left-top, left-bottom, right-top, or right-bottom. This positional information is then incorporated into a new location prompt (e.g., “The monopolar curved scissors on the right bottom”), which serves as input for the second inference. The model generates the segmentation map \(M_c^3\) and the existence probability \(p_c^3\) for the class \(c\).

For the final output \(M_c\), the segmentation maps from both iterations are combined using as below:
\begin{align}
\label{eq:eq3}
    M_c=
    \begin{cases}
    0, & \text{if}\ \frac{p_c^1 + p_c^2}{2} < 0.5\\
    \frac{M_c^1 + M_c^2}{2}, &  \frac{p_c^1 + p_c^2}{2}  \geq 0.5\ \& \ p_c^3 < 0.5 \\
    \frac{M_c^1 + M_c^2 + M_c^3}{3}, & \frac{p_c^1 + p_c^2}{2}  \geq 0.5\ \& \ p_c^3 \geq 0.5
    \end{cases}
\end{align}

This combination leverages both the fixed prompts and the location prompt, resulting in a more precise and contextually aware segmentation map. This iterative refinement approach ensures robust segmentation by progressively refining predictions based on both textual and spatial prompts, enhancing segmentation performance in complex surgical images. We describe the effects of iterative refinement in detail in the ablation study in Table \ref{tab:tab4}.

\begin{table}[t]
\centering
\caption{Prompt information for each class generated from GPT-4\cite{achiam2023gpt}.}
\begin{tabular}{@{}p{3.5cm}p{12.5cm}@{}}
\toprule
\multicolumn{2}{l}{\textbf{GPT-4 prompt information}} \\ \midrule
Question                  & Please describe the appearance of the \{cls name\} in endoscopic surgery with 18 words, change the description to a phrase with the subject, and do not use colons and ``endoscopic surgery" and ``minimally invasive procedure". \\ \midrule
\textbf{Class name}                & \textbf{Answer} \\ \midrule
Bipolar forceps           & Bipolar forceps feature slim insulated handles and precise tips for targeted tissue coagulation and manipulation.                                                                                                               \\
Prograsp forceps          & Prograsp forceps feature ergonomic handles and curved serrated jaws for secure and precise tissue manipulation during surgery.                                                                                                  \\
Large needle drivers & Large needle drivers are equipped with sturdy handles and sharp precise jaws for efficient suturing in surgical settings.                                                                                                       \\
Monopolar curved scissors & Monopolar curved scissors have slender ergonomic handles and sharp curved blades for precise cutting during surgical operations.                                                                                                \\
Ultrasound probe          & The ultrasound probe features a slim elongated design with a smooth tip for detailed imaging during surgical interventions.                                                                                                     \\
Suction instrument        & The suction instrument boasts a narrow elongated tube with a controllable tip for precise fluid removal during surgery.                                                                                                         \\
Clip applier              & The clip applier has a long slender shaft with a specialized tip for deploying clips securely during surgical procedures.                                                                                                       \\
Vessel sealer             & Vessel sealer features a streamlined handle and specialized tips for precise sealing and division of blood vessels during surgery.                                                                                              \\
Grasping retractor        & Grasping retractor sports a long handle with claw-like tips for firmly holding tissues during surgical operations.                                                                                                              \\ \bottomrule
\end{tabular}
\label{t1}
\end{table}

\section{Experiment}
\subsection{Dataset and metric}
We evaluate our method using the EndoVis2017 \cite{allan20192017} and EndoVis2018 \cite{allan20202018} datasets based on endoscopic surgery videos. The EndoVis2017 dataset consists of 10 videos recorded using the da Vinci surgical robot system, each containing 255 frames. This dataset includes seven classes: bipolar forceps (BF), prograsp forceps (PF), large needle driver (LND), vessel sealer (VS), grasping retractor (GR), monopolar curved scissors (MCS), and ultrasound probe (UP). Following \cite{shvets2018automatic}, we apply 4-fold cross-validation to assess performance. The EndoVis2018 dataset consists of 15 videos, divided into 11 training and 4 test video sequences, and includes the following seven classes: bipolar forceps, prograsp forceps, large needle driver, monopolar curved scissors, ultrasound probe, suction instrument (SI), and clip applier (CA). The EndoVis2018 dataset exhibits severe class imbalance in the training set, and to ensure balanced learning, we adjust class frequencies within the dataloader. 

As mentioned in Section \ref{sec:sec3.2}, we use three fixed prompts during training for each class. The first prompt incorporates the class name, i.e., ``The \{cls name\},'' the second refers to the instrument's location, i.e., ``The \{cls name\} on the \{location\},'' and the third prompt is generated from GPT-4. 
As shown in Table \ref{t1}, the prompts generated by GPT-4 describe the appearance of the class based on its name. During inference, we only use the class name and the GPT-4 prompts, as the location information is not available at the start.

We follow the evaluation metrics used in previous works \cite{gonzalez2020isinet, zhou2023text}, which include Challenge IoU (Ch IoU), ISI IoU, and mean class IoU (mc IoU). Ch IoU calculates the IoU for classes present in the image and then computes the average. ISI IoU measures the IoU across all classes, while mc IoU represents the average IoU for each class. 
\begin{table}[ht]
\centering
\caption{Comparison of previous methods in terms of domain, input data type, and robustness evaluation.}
\begin{tabular}{@{}l|c|c|c}
\toprule
Method                                & Domain         & Input & Robustness  \\ \midrule
TernausNet-11 \cite{iglovikov2018ternausnet} & vision  & I & clear  \\
TraSeTR \cite{zhao2022trasetr}        & vision         & S & clear   \\
MF-TAPNet \cite{jin2019incorporating} & vision         & S & clear   \\
Dual-MF \cite{zhao2020learning}       & vision         & S & clear                          \\
ISINet \cite{gonzalez2020isinet}      & vision         & S & clear                          \\
MATIS \cite{ayobi2023matis}           & vision         & S & data leakage             \\
S3Net \cite{baby2023forks}            & vision         & I & use only top 5 instance           \\ \midrule
TP-SIS \cite{zhou2023text}            & promptable     & I & oracle information \\
RoSIS (ours)                          & promptable     & I & clear         \\ \bottomrule                
\end{tabular}
\label{table:table1}
\end{table}

\subsection{Implementation detail}
We implement our model using PyTorch \cite{paszke2019pytorch}. For the language model, we utilize a pre-trained BERT \cite{devlin2018bert}. The image encoder is initialized with an ImageNet \cite{deng2009imagenet} pre-trained Swin Transformer \cite{liu2021swin} to support dense prediction. In our fusion module, we set the language token size to 20. The model is trained using the AdamW \cite{loshchilov2017decoupled} optimizer with a weight decay of 0.01 for 50 epochs, applying a learning rate decay of 0.1 at epochs 30 and 40. The input images, originally sized at 1280$\times$1024, are resized to 480$\times$480 for training, and the generated masks are upsampled to their original size to produce the final mask. We train the model with a batch size of 32, using four Nvidia RTX A6000 GPUs.

\subsection{Comparative results}
Table \ref{table:table1} presents a comparison of previous methods in terms of domain, input data type, and robustness evaluation. The domain categorization differentiates between vision-based methods, which rely solely on image features, and promptable methods, which incorporate additional guidance such as textual prompts for segmentation. The input data type column distinguishes methods that operate on a single image (I) from those that leverage a sequence of images (S) to enhance segmentation performance using temporal information. The robustness column indicates whether each method has been evaluated under robust conditions. Methods such as MATIS \cite{ayobi2023matis}, S3Net \cite{baby2023forks}, and TP-SIS \cite{zhou2023text} do not meet robustness evaluation criteria due to issues like data leakage or oracle information. We exclude MATIS from the comparison methods because it conducts experiments excluding certain videos during the experiment. S3Net uses whole train/test data, but it selects the top 5 instances out of 7 instances during inference. TP-SIS is re-implemented under robust conditions and used in experiments.

In Tables \ref{tab:tab1} and \ref{tab:tab2}, we compare our method, RoSIS, with both vision-based and promptable models on the EndoVis2018 \cite{allan20202018} and EndoVis 2017 \cite{allan20192017} datasets. For a fair comparison, we re-implement the promptable models LAVT \cite{yang2022lavt} and TP-SIS \cite{zhou2023text}, both of which are based on referring image segmentation tasks. For promptable methods, we provide referring sentences for all classes in the dataset. Pixel-wise probabilities of the generated masks were then compared to produce the final output. 

\begin{table}[t]
\centering
\footnotesize
\caption{Comparison results on the EndoVis2018 dataset. * means re-implemented results. $\dagger$ means inference under not robustness conditions. \textbf{Bold} and \uline{underline} indicate the best and the second best performance, respectively.}
\begin{tabular}{c|l|ccc|ccccccc}
\toprule

\multirow{2}{*}{Domain} & \multirow{2}{*}{Method} & \multirow{2}{*}{Ch IoU} & \multirow{2}{*}{ISI IoU} & \multirow{2}{*}{mc IoU} & \multicolumn{7}{c}{Instrument category} \\
& & & & & BF    & PF    & LND   & SI          & CA    & MCS            & UP    \\ \midrule
\multirow{5}{*}{Vision} & TernausNet-11 \cite{iglovikov2018ternausnet} & 46.22 & 39.87 & 14.19 & 44.20 & 4.67 & 0.00 & 0.00 & 0.00 & 50.44 & 0.00 \\
        & TraSeTR \cite{zhao2022trasetr} & 76.20 & - & \textbf{47.77} & 76.30 & \textbf{53.30} & \textbf{46.50} & 40.60 & \textbf{13.90} & 86.30 & \textbf{17.50} \\
        & MF-TAPNet \cite{jin2019incorporating} & 67.87 & 39.14 & 24.68 & 69.23 & 6.10 & 11.68 & 14.00 & 0.91 & 70.24 & 0.57 \\
        & Dual-MF \cite{zhao2020learning}  & 70.40 & - & 35.09 & 74.10 & 6.80 & \uline{46.00} & 30.10 & \uline{7.60} & 80.90 & 0.00 \\
        & ISINet \cite{gonzalez2020isinet} & 73.03 & 70.97 & 40.21 & 73.83 & 48.61 & 30.98 & 37.68 & 0.00 & \uline{88.16} & 2.16 \\
        & S3Net$^\dagger$ \cite{baby2023forks} & 75.81 & \uline{74.02} & 42.58 & 77.22 & \uline{50.87} & 19.83 & \textbf{50.59} & 0.00 & \textbf{92.12} & 7.44 \\ \midrule
\multirow{4}{*}{Promptable} & LAVT$^*$ \cite{yang2022lavt}  & 72.87 & 32.58 & 25.97 & 77.19 & 10.96 & 8.79 & 16.47 & 2.78 & 64.04 & 1.55 \\
        & RefSegformer \cite{wu2024towards}  & \uline{77.09} & 68.67 & 36.70 & \uline{80.67} & 20.88 & 25.06 & 38.86 & 0.10 & 86.23 & 5.13 \\
        & TP-SIS$^*$ \cite{zhou2023text}  & 70.55 & 55.37 & 29.93 & 72.12 & 12.33 & 15.66 & 20.56 & 1.98 & 85.56 & 1.30 \\
        & RoSIS (ours) & \textbf{77.66} & \textbf{76.16} & \uline{44.54} & \textbf{84.11} & 42.73 & \uline{35.26} & 46.36 & 0.00 & 88.10 & \uline{15.22} \\ \bottomrule
\end{tabular}
\label{tab:tab1}
\end{table}

\begin{table}[ht]
\centering
\footnotesize
\caption{Comparison results on the EndoVis2017 dataset. * means re-implemented results. $\dagger$ means inference under not robustness conditions. \textbf{Bold} and \uline{underline} indicate the best and the second best performance, respectively.}
\begin{tabular}{c|l|ccc|ccccccc}
\toprule

\multirow{2}{*}{Domain} & \multirow{2}{*}{Method} & \multirow{2}{*}{Ch IoU} & \multirow{2}{*}{ISI IoU} & \multirow{2}{*}{mc IoU} & \multicolumn{7}{c}{Instrument category} \\
& & & & & BF    & PF    & LND   & VS          & GR    & MCS            & UP    \\ \midrule
\multirow{5}{*}{Vision} & TernausNet-11 \cite{iglovikov2018ternausnet} & 35.27 & 12.67 & 10.17 & 13.45 & 12.37 & 20.51 & 5.97 & 1.08 & 1.00 & 16.76 \\
& TraSeTR \cite{zhao2022trasetr}  & 60.40 & - & 36.79 & 45.20 & \textbf{56.70} & 55.80 & 38.90 & \uline{11.40} & 31.30 & 18.20 \\
        & MF-TAPNet \cite{jin2019incorporating} & 37.35 & 13.49 & 10.77 & 16.39 & 14.11 & 19.01 & 8.11 & 0.31 & 4.09 & 13.40 \\
        & Dual-MF \cite{zhao2020learning}  & 45.80 & - & 26.40 & 34.40 & 21.50 & \textbf{64.30} & 24.10 & 0.80 & 17.90 & 21.80 \\
        & ISINet \cite{gonzalez2020isinet}  & 55.62 & 52.20 & 28.96 & 38.70 & 38.50 & 50.09 & 27.43 & 2.01 & 28.72 & 12.56 \\
        & S3Net$^\dagger$\cite{baby2023forks}  & \textbf{72.54} & \textbf{71.99} & \textbf{46.55} & \textbf{75.08} & \uline{54.32} & \uline{61.84} & 35.50 & \textbf{27.47} & \uline{43.23} & 28.38 \\ \midrule
\multirow{4}{*}{Promptable}  & LAVT$^*$ \cite{yang2022lavt}  & 37.57 & 11.10 & 10.90 & 17.05 & 12.97 & 14.25 & 11.91 & 2.23 & 7.98 & 9.90 \\
& RefSegformer \cite{wu2024towards}  & 46.58 & 32.81 & 27.87 & 33.37 & 0.30 & 52.69 & 38.50 & 0.03 & 37.05 & \uline{33.14} \\
 & TP-SIS$^*$ \cite{zhou2023text}  & 51.36 & 46.64 & 31.24 & 43.81 & 27.70 & 46.42 & \uline{41.25} & 3.98 & \textbf{46.83} & 8.70 \\
        & RoSIS (ours)  & \uline{63.07} & \uline{57.33} & \uline{40.65} & \uline{67.34} & 39.95 & 55.85 & \textbf{45.32} & 5.10 & 26.00 & \textbf{44.99} \\ \bottomrule
\end{tabular}
\label{tab:tab2}
\end{table}

\subsubsection{Comparison on the EndoVis2018}
As shown in Table \ref{tab:tab1}, RoSIS outperforms all promptable segmentation models and achieves superior performance over vision-based methods in Ch IoU, ISI IoU, and mc IoU. S3Net is the strongest vision-based model in ISI IoU that utilizes a single image. However, RoSIS surpasses it under robust conditions, improving Ch IoU by +2.44\%, ISI IoU by +2.89\%, and mc IoU by +4.60\%. Meanwhile, TraSeTR, which achieves the highest mc IoU among vision-based models, benefits from using sequential images instead of single-image inference. Despite this advantage, RoSIS remains competitive while relying solely on single-image processing.

Compared to promptable models like LAVT and TP-SIS, RoSIS effectively reduces false positives and false negatives. Existing promptable methods generate segmentation masks without first verifying instrument existence, causing a noticeable gap between Ch IoU and ISI IoU scores. In contrast, RoSIS incorporates an existence verification step, stabilizing the segmentation process by minimizing false negatives. This refinement results in higher segmentation accuracy, making RoSIS a more reliable solution for robust surgical instrument segmentation.

\subsubsection{Comparison on the EndoVis2017}
Table~\ref{tab:tab2} presents a detailed comparison of segmentation performance on the EndoVis2017 \cite{allan20192017} dataset. Among the vision-based models, S3Net \cite{baby2023forks} achieves the highest overall performance. However, this performance is achieved under a setting that retains only the top five predicted instances per frame, which may simplify the segmentation task by limiting the number of considered categories. While this assumption aligns with common surgical scenes where fewer than five instruments are present, it reduces flexibility when applied to scenarios involving a broader set of instruments or multi-phase procedures.

Moreover, RoSIS outperforms other vision-based and promptable models while maintaining a more balanced class-wise performance distribution. Notably, its ISI IoU shows less performance degradation compared to Ch IoU, highlighting the effectiveness of its fusion structure, prompt design, and iterative refinement strategy. Under robust conditions, except for S3Net, RoSIS shows competitive performance overall and achieves the best performance in VS and UP. These results further validate RoSIS’s ability to effectively integrate vision-based and promptable features, making it a versatile and efficient solution for surgical instrument segmentation.

\begin{table}[t]
\centering
\caption{Comparison of promptable segmentation models under the R-SIS setting using FPR, Precision, Recall, and F1-score. \textbf{Bold} and \uline{underline} indicate the best and the second best performance, respectively.}
\label{tab:model_fpr_precision_recall}
\begin{tabular}{l|l|cccc}
\toprule
Dataset & Method & FPR ↓ & P ↑ & R ↑ & F1 ↑\\
\midrule
\multirow{5}{*}{EV18} & LAVT \cite{yang2022lavt}        & 0.6155 & 0.3770 & 0.3178 & 0.3364 \\
& RefSegformer \cite{wu2024towards} & 0.2688 & 0.6820 & 0.6369 & 0.6420 \\
& TP-SIS \cite{zhou2023text}      & 0.4008 & 0.5374 & 0.4866 & 0.4957 \\
& RoSIS w/o exist & \uline{0.1804} & \uline{0.7374} & \uline{0.6714} & \uline{0.6847} \\
& RoSIS (Ours) & \textbf{0.0892} & \textbf{0.7926} & \textbf{0.7546} & \textbf{0.7611} \\
\midrule
\multirow{5}{*}{EV17} & LAVT \cite{yang2022lavt}        & 0.7744 & 0.2153 & 0.1351 & 0.1483 \\
& RefSegformer \cite{wu2024towards} & 0.4706 & 0.3505 & 0.2578 & 0.2732 \\
& TP-SIS \cite{zhou2023text}      & \uline{0.3338} & 0.4802 & 0.3942 & 0.3981 \\
& RoSIS w/o exist & 0.3532 & \uline{0.5445} & \uline{0.4427} & \uline{0.4596} \\
& RoSIS (Ours) & \textbf{0.2343} & \textbf{0.5907} & \textbf{0.5036} & \textbf{0.5269} \\
\bottomrule
\end{tabular}
\end{table}

\subsubsection{Robustness-Oriented Comparison}
Table~\ref{tab:model_fpr_precision_recall} presents a comparative analysis of segmentation models under the R-SIS setting on the EndoVis2017 and EndoVis2018 datasets. We compare the performance of segmentation models under the R-SIS setting using false positive rate (FPR), precision (P), recall (R), and F1 score (F1). The results show that RoSIS substantially reduces the occurrence of false-positive predictions compared to prior promptable segmentation models. While existing methods often produce masks for absent instruments due to the assumption of instrument presence in prompts, RoSIS effectively suppresses these errors by incorporating an existence prediction mechanism. In addition to its low FPR, RoSIS achieves consistently high precision and recall, demonstrating that its robustness does not come at the cost of under-segmentation. These findings confirm the advantage of the R-SIS framework and support the use of prompt-agnostic prompting strategies for fair and reliable evaluation. Table~\ref{tab:model_fpr_precision_recall} also introduces results for RoSIS without the existence prediction module (w/o exist), allowing direct comparison the effectiveness of this component. The existence prediction module leads to a clear reduction in false positives and a noticeable gain in precision across both datasets, confirming its contribution to segmentation robustness. Notably, even without the existence prediction module, RoSIS outperforms prior promptable segmentation models, indicating that its core design, rooted in robust prompting strategies and R-SIS based task framing, is inherently better suited to the evaluation conditions.

\begin{figure}[t]
    \centering
    \includegraphics[width=0.7\linewidth]{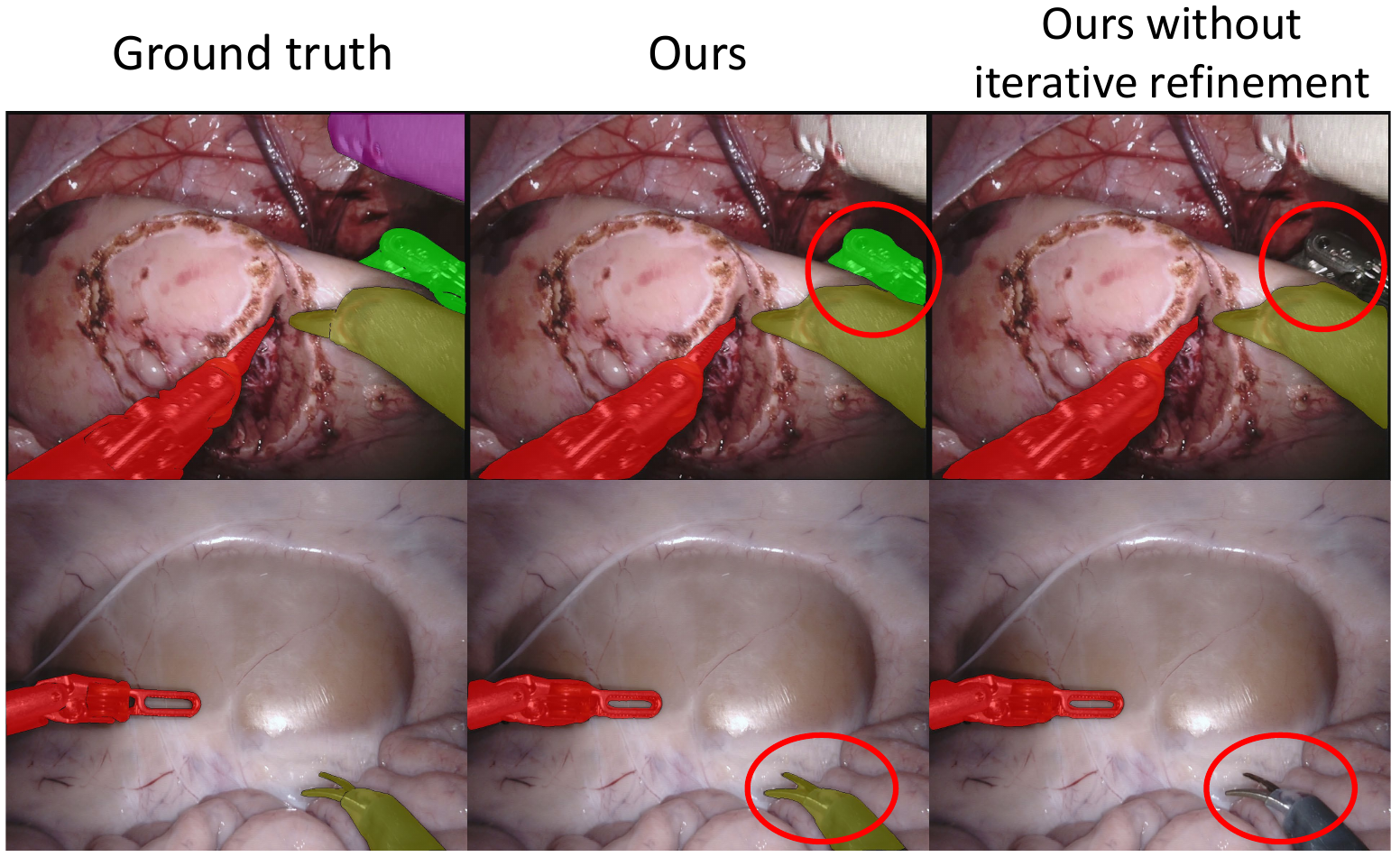}
    \caption{Visual comparison of iterative refinement strategies}
    \label{fig:fig5}
\end{figure}

\begin{table}[t]
\centering
\caption{Text prompt design and iterative refinement ablation study in EndoVis2018 dataset. The values in brackets represent the increase compared to the baseline.}
\begin{tabular}{l|ccc}
\toprule
                                & ISI IoU & Ch IoU & mc IoU \\ \midrule
Name prompt                     & 67.74 & 76.20 & 37.84  \\
(+) location prompt       & 69.24 & 76.57 & 37.54  \\
\ \ \ (+) iterative refinement & 73.51  & 77.58 & 39.30  \\
(+) GPT-4 prompt               & 71.37  & 77.21 & 40.61  \\
\ \ \ (+) iterative refinement & \textbf{76.16} & \textbf{77.66} & \textbf{44.54} \\ \bottomrule
\end{tabular}
\label{tab:tab4}
\end{table}

\subsection{Ablation study}
\subsubsection{Iterative refinement process and prompt design}
Table \ref{tab:tab4} presents the ablation study results on the EndoVis2018 dataset, emphasizing the role of iterative refinement and text prompts. Starting with the baseline using only the name prompt such as “The bipolar forceps,” the model achieves ISI IoU and mc IoU scores of 67.74 and 37.84, respectively. Adding the location prompt to the name prompt slightly improves performance in ISI IoU. Then, the iterative refinement consistently improves performance across all metrics. This demonstrates that iterative refinement effectively refines the segmentation map by utilizing location-specific prompts, contributing to more accurate segmentation. Similarly, when iterative refinement is incorporated with the GPT-4 prompt, which provides descriptive characteristics for each instrument, there is a substantial improvement in segmentation accuracy. The ISI IoU increases by 4.79, and mc IoU increases by 3.93. This improvement shows the synergy between descriptive prompts and iterative refinement, as the iterative process allows the model to refine segmentation based on more detailed instrument information. Also, as shown in the red circle part of Fig. \ref{fig:fig5}, iterative refinement allows for more accurate segmentation maps.

These results indicate the crucial role of iterative refinement in our model. By refining segmentation through multiple stages and adapting prompts based on previous outputs, iterative refinement contributes to significant gains in both general and class-specific accuracy, making it a key factor in achieving optimal segmentation performance.
\begin{table}[t]
\centering
\caption{A visual and language feature fusion structural ablation study and performance comparison of the fusion structure on the EndoVis2018 dataset.}
\begin{tabular}{l|cccc|ccc}
\toprule
Method & MMFB & SGB & RL & LT &  ISI IoU & Ch IoU  & mc IoU    \\     \midrule
 \multirow{4}{*}{Ours}& \cmark & \xmark & \xmark & \xmark & 64.31 & 72.75 & 34.27   \\
                      & \cmark & \cmark & \xmark & \xmark & 71.65 & 77.07 & 38.23   \\
                      & \cmark & \cmark & \cmark & \xmark & 74.36 & \textbf{80.37} & 41.87   \\
                      & \cmark & \cmark & \cmark & \cmark& \textbf{76.16} & 77.66 & \textbf{44.54}  \\ \midrule
\multicolumn{5}{l|}{VLTF (RefSegformer) \cite{wu2024towards}} &68.58 & 74.19        & 33.48          \\
\multicolumn{5}{l|}{PWAM (LAVT) \cite{yang2022lavt}} &66.94 & 79.34    & 36.89          \\ \bottomrule
\end{tabular}
\label{tab:tab3}
\end{table}

\subsubsection{Structure design}
To evaluate the effectiveness of our visual-language fusion structure, we conduct an ablation study on the EndoVis2018 dataset, analyzing the impact of each module on segmentation performance. Specifically, we assess the contributions of the Multi-Modal Fusion Block (MMFB), Selective Gate Block (SGB), Raw Language features (RL), and Language Token (LT) by incrementally adding components to our framework. RL involves using language features extracted from BERT \cite{devlin2018bert} directly as input for the MHCA module in the existence prediction branch, enhancing textual representation within our segmentation framework. The results are summarized in Table \ref{tab:tab3}.

Our findings demonstrate that each proposed module contributes to segmentation performance. SGB improves segmentation stability by selectively refining the fusion of visual and language features. RL enhances the model’s ability to process textual information, leading to slight improvements in segmentation accuracy. LT further strengthens feature representation, helping the model to refine segmentation masks more effectively. When combined, these modules achieve the best overall performance, validating their role in improving segmentation robustness.

To evaluate the effectiveness of our fusion structure, we compare our model with previous RIS methods, including VLTF (RefSegformer) \cite{wu2024towards} and PWAM (LAVT) \cite{yang2022lavt}. The results show that our method improves segmentation accuracy and maintains stable performance. Unlike previous methods that rely heavily on either vision or language cues, our approach balances multi-modal information more effectively, reducing misclassification and false positives. The consistent improvement across key metrics confirms its robustness and reliability in surgical instrument segmentation.

\begin{figure}[t]
    \centering
    \includegraphics[width=\textwidth]{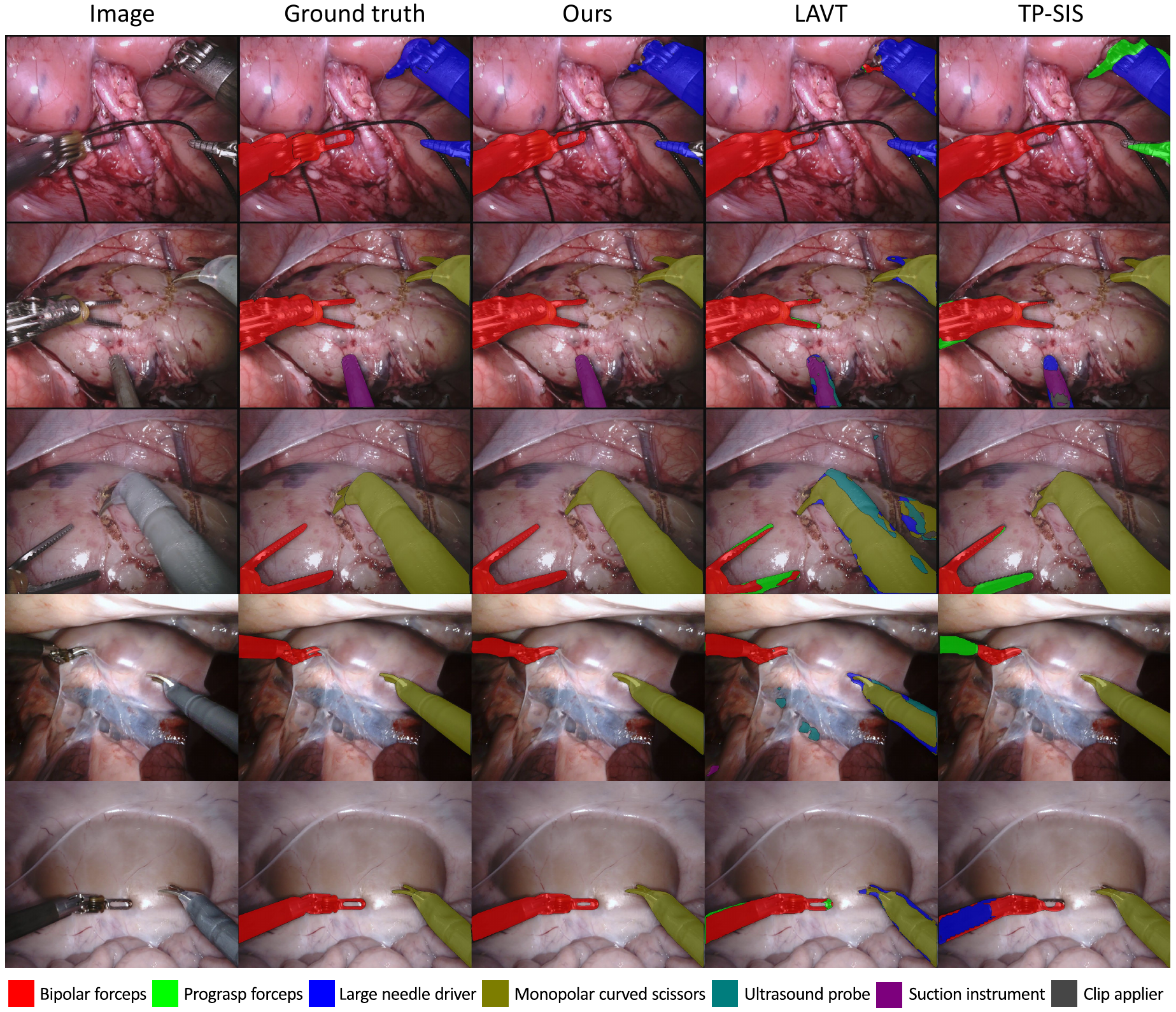}
    \caption{Qualitative results of our predicted maps, the ground truth masks, and other promptable segmentation methods from the EndoVis2018 validation set.}
    \label{fig:fig6}
\end{figure}

\subsection{Qualitative results}
In Fig. \ref{fig:fig6}, we visualize the prediction of our model, the groud truth, and the other promptable segmentation methods \cite{yang2022lavt, zhou2023text}. Since other methods are trained without robust conditions, they hard to determine the presence or absence of instruments in the R-SIS environment, resulting in higher false positives in the predicted maps. As shown in the 4th and 5th columns of Fig. \ref{fig:fig6}, the segmentation masks for various classes overlap. This overlap occurs because segmentation maps are generated for all classes' text prompts, causing all maps to merge. In contrast, our model first verifies the presence or absence of an instrument before generating a map, producing a clean segmentation for a single instrument and leading to a low false positive rate, which is crucial in surgical environments.

\begin{table}[t]
\centering
\caption{Cross-dataset setting performance comparison by training on the EndoVis2018 and testing on the EndoVis2017, and vice versa.}
\begin{tabular}{c|l|ccc}
\toprule
Setting & Method      & Ch IoU & ISI IoU & mc IoU  \\ \midrule
\multirow{3}{*}{EV18$\rightarrow$ EV17} & LAVT \cite{yang2022lavt}  & 40.82 & 14.35 & 13.26 \\
& TP-SIS \cite{zhou2023text} & 34.23 & 18.57 & 20.16 \\
& RoSIS (ours)   & \textbf{42.45} & \textbf{34.67} & \textbf{27.68} \\ \midrule
\multirow{3}{*}{EV17$\rightarrow$EV18} & LAVT \cite{yang2022lavt} & 46.57 & 15.67 & 14.91 \\
&TP-SIS \cite{zhou2023text} & 37.81 & 26.72 & 14.93 \\
&RoSIS (ours)   & \textbf{58.31} &\textbf{51.00} & \textbf{23.99} \\ \bottomrule
\end{tabular}
\label{tab:tab5}
\end{table}

\subsection{Cross-dataset study}
We also evaluate our method in a cross-dataset setting. In this setup, the model is trained on the EndoVis2018 dataset and tested on the EndoVis2017 dataset, and vice versa. The EndoVis2017 and EndoVis2018 datasets share five common categories and each includes two unique categories. As shown in Table \ref{tab:tab5}, RoSIS demonstrates better performance compared to other promptable segmentation methods. Additionally, even among shared categories, there are slight differences in shape and domain variations, making the cross-dataset experiment indicate the generalizability of methods.

The promptable segmentation methods LAVT \cite{yang2022lavt} and TP-SIS \cite{zhou2023text} show overall lower performance, with a large discrepancy between their Ch IoU and ISI IoU scores. In contrast, our method, RoSIS, achieves the best results, with a smaller gap between Ch IoU and ISI IoU, showing its robustness and efficiency. These results demonstrate the generalizability and effectiveness of RoSIS in handling domain shifts across different datasets.

\section{Conclusion}
In this work, we redefined the task of text-promptable Surgical Instrument Segmentation (SIS) under robust conditions as Robust text-promptable SIS (R-SIS). Existing methods often assume the presence of instruments described by prompts, leading to challenges in realistic surgical scenarios. Such assumptions introduce biases in evaluation, as they leverage prior knowledge that is not available to vision-based models. R-SIS eliminates this dependency by enforcing prompt-based segmentation without prior knowledge of instrument presence. This ensures fairer comparisons between promptable and vision-based methods while improving robustness in real-world surgical applications.

To support this task setting, we developed Robust Surgical Instrument Segmentation (RoSIS), effective framework designed to address prompt uncertainty and instrument absence. Rather than relying on architectural complexity, RoSIS integrates prompt types with an existence prediction mechanism and an iterative refinement strategy, enabling reliable segmentation under R-SIS constraints.

Our experiments on the EndoVis2017 and EndoVis2018 datasets demonstrate that RoSIS consistently reduces false positive predictions while maintaining strong precision and recall. Ablation studies further validate that even without existence prediction, RoSIS performs competitively against existing models, underscoring the value of our task-aware formulation. This work highlights the importance of fair evaluation protocols for promptable segmentation and provides a foundation for future research in multi-modal surgical scene understanding.

\section{Acknowledgements}
This work was supported by the Technology Innovation Program (RS-2024-00443054, Development of a Supermicrosurgical Robot System for Sub-0.8mm Vessel Anastomosis through Human-Robot Autonomous Collaboration in Surgical Workflow Recognition) funded By the Ministry of Trade Industry \& Energy (MOTIE, Korea).







\end{document}